\RequirePackage{amsmath,amssymb}
\documentclass{template/pkuai4m}
\microtypesetup{expansion=false}

\usepackage[utf8]{inputenc}
\usepackage[T1]{fontenc}
\usepackage{enumitem}
\setlist[itemize]{leftmargin=*, itemsep=0.2em, topsep=0.3em, beginpenalty=10000}
\usepackage{listings}
\lstdefinestyle{leanresult}{
  basicstyle=\ttfamily\footnotesize,
  columns=fullflexible,
  keepspaces=true,
  mathescape=true,
  showstringspaces=false,
  breaklines=true,
  frame=single,
  rulecolor=\color{black!25},
  backgroundcolor=\color{black!3},
  framesep=6pt,
  xleftmargin=6pt,
  xrightmargin=6pt,
  literate={ℝ}{{$\mathbb{R}$}}1
           {∀}{{$\forall$}}1 {∞}{{$\infty$}}1
           {≃ₜ}{{$\simeq_{\mathrm{t}}$}}2
}

\title{An AI-Assisted Formalization of the Poincar\'e Conjecture}

\renewcommand{\affiliation}[2][]{%
  \addtolist[#1]{#2}{\affiliationlist}{\affiliationformat}{\\}}

\author[1,*]{Zhiyuan Zhang}
\author[2,*]{Axel Delaval}
\author[2,8]{Leheng Chen}
\author[2]{Jinxuan Chen}
\author[2]{Jie Xu}
\author[3]{Yuxuan Liao}
\author[1]{Jiedong Jiang}
\author[4,\P]{Chunlei Liu}
\author[5,6,7,8,\P]{Bin Dong}

\affiliation[1]{School of Mathematical Sciences, Peking University}
\affiliation[2]{Beijing International Center for Mathematical Research, Peking University}
\affiliation[3]{School of Mathematical Sciences, Beijing Normal University}
\affiliation[4]{School of Mathematical Sciences, Capital Normal University}
\affiliation[5]{Beijing International Center for Mathematical Research and the New Cornerstone Science Laboratory, Peking University}
\affiliation[6]{Center for Machine Learning Research, Peking University}
\affiliation[7]{Center for Intelligent Computing, Great Bay Institute for Advanced Study, Great Bay University}
\affiliation[8]{Zhongguancun Academy}

\contribution[*]{Equal contribution}
\contribution[\P]{Corresponding author}
\def\emailicon{\raisebox{-1.5pt}{\includegraphics[height=1.05em]{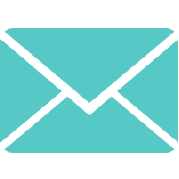}}}
\def\githubicon{\raisebox{-1.5pt}{\includegraphics[height=1.05em]{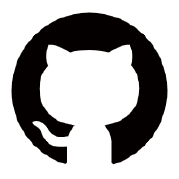}}}
\checkdata[\emailicon\hspace{0.3em} Correspondence]{\email{7357@cnu.edu.cn}, \email{dongbin@math.pku.edu.cn}}
\checkdata[\githubicon\hspace{0.3em} Repository]{\url{https://github.com/frenzymath/PoincareConjecture}}
\checkdata[Website]{\url{https://frenzymath.github.io/Poincare-Conjecture/\#/}}

\abstract{We present an AI-assisted Lean 4 formalization of the Poincar\'e conjecture. The project began with limited reusable formal infrastructure for the geometric analysis behind the proof. To organize this work, we combined a proof blueprint prepared by mathematicians with explicit milestone statements. These milestones enabled parallel agent work and gave mathematicians clear points to locate blockers and provide effective mathematical guidance. Our analysis identifies the human interventions and organizational choices behind this workflow. The project provides a starting point toward reusable infrastructure for future formalization projects; such infrastructure, once developed, could eventually reduce the cost of verifying mathematical results in geometric analysis.}

\begin{document}
\maketitle

\section{Introduction}
\label{sec:introduction}

Checking a proof is one of the most common and demanding tasks in mathematics, and it becomes especially difficult when the argument spans several fields. Recently, AI has greatly accelerated the production of proofs, making reliable verification ever more important. Proof assistants such as Lean~4~\cite{lean4} verify proofs by rewriting a mathematical argument into precise formal code that the computer mechanically checks, thereby ensuring soundness. Although such systems have incorporated automation to reduce the burden of formalization, constructing formal code still requires extensive manual effort and expertise. Recent advances have enabled large language models to formalize, reducing this manual burden, and only very recently have they become capable of doing so at scale. Once a generated formal proof has passed kernel checking under the permitted axioms, the remaining human task is to judge whether its statement, assumptions, and definitions are semantically faithful to the original mathematical content; a reviewer can then verify the result without reconstructing every proof step or mastering every technique used to establish it. Still, expert guidance can be of great help throughout the formalization process in choosing definitions, reviewing statements, assessing code organization, and resolving mathematical difficulties during proof construction.

Early work on autoformalization studied the translation of mathematical statements~\cite{wu2022autoformalization,jiang2023mma,gao2025herald}, alongside progress in generating formal proofs~\cite{polu2020gptf,yang2023leandojo,ren2025deepseek}. As language models have become increasingly capable of producing proofs in natural language, the bottleneck has shifted from constructing a proof to translating it into formal code. This shift has enabled an end-to-end pipeline spanning generation, translation, and formal verification, which agent systems sustain through library retrieval, compiler feedback, and repeated task execution~\cite{thakur2024copra,liu2026numina,ju2026archon}. In parallel, the scale of projects that autoformalization can handle has grown from individual proofs to bodies of interrelated results and textbook-level theories~\cite{urban2026topology,wang2026m2f,gloeckle2026textbook}. The September 2026 formalization of Fermat's Last Theorem using Prove2Me produced a complete computer-checked proof of unprecedented scale~\cite{prove2me2026,anthropicFLT2026}, but how such large-scale formalizations are actually organized, and where human and AI effort is divided, is not yet widely understood.

We use the formalization of the Poincar\'e conjecture as a testing ground for these questions, a project whose scale and scope make it unusually demanding for organizing human and AI effort. The conjecture, proved by Perelman, states that every simply connected, closed three-manifold is homeomorphic to the three-sphere. We follow Morgan--Tian's detailed exposition~\cite{MT2007}. The proof requires geometric analysis for which Mathlib still lacks mature reusable interfaces, making the building of foundations an essential part of the project.

Our preparation began in May 2026 with three AI engineers with mathematical training, who worked alongside a group of mathematicians. Over the following months, we developed a natural-language blueprint of the proof and formalized some background material. Drawing on this preparation, we substantially redesigned our workflow
and restarted the main formalization in September with about 90 milestones,
each specifying the definitions and Lean statement required for its target. The milestones provided shared mathematical targets, exposed dependencies, and made
progress and blockers visible. The engineers used this structure to
coordinate parallel agent work, while mathematicians reviewed the final
theorem and key intermediate statements and helped resolve mathematical
difficulties during proof construction.

Using publicly available commercial models, we completed the main
formalization phase in slightly more than two weeks, after the initial preparation phase. AI subscriptions and server rentals cost about \$25{,}000.
The resulting development proves both the topological and smooth
Poincar\'e theorems.\footnote{
An independent formalization of the Poincar\'e conjecture was also
announced by Ziyang Qin, Yuan Liao, Bennett Chow, Ayush Khaitan, and
Jack McCarthy~\cite{differentialGeometry2026}.}
It passes a full Lean build and Comparator verification against target
statements written using only Mathlib.\footnote{
The \href{https://github.com/frenzymath/PoincareConjecture}{repository}
includes the independent
\href{https://github.com/frenzymath/PoincareConjecture/blob/main/Comparator/Challenge.lean}{target statements}
and verification instructions.
\href{https://github.com/leanprover/comparator}{Comparator} checks statement
agreement and permitted axioms; our configuration also checks the proofs
with the independent Nanoda kernel.}
The initial development contained about $3.2$ million lines of Lean code.\footnote{Both figures count Lean source lines after removing comments, using the versions before and after \href{https://github.com/frenzymath/PoincareConjecture/pull/3}{PR~\#3}, respectively.}
Extracting the dependencies of the final theorem reduced it to about
$2.8$ million lines while preserving verification.

Section~\ref{sec:background} reviews the mathematical background and
related work. Section~\ref{sec:methodology} presents the workflow, team
organization, and project resources. Section~\ref{sec:analysis} examines
agent behavior, mathematical interventions, and the sources of
acceleration, and compares our development with the FLT formalization.
Section~\ref{sec:conclusion} discusses the prospects for reusable
mathematical infrastructure and future formalization projects.

\section{Background and Related Work}
\label{sec:background}

\paragraph{Lean and Mathlib.}
Lean~4 is a programming language and proof assistant in which mathematics is formulated as code: definitions, theorem statements, and proofs are all written in Lean, and compiling the code checks each proof against its statement in Lean's kernel~\cite{lean4}. A proof that passes kernel checking is guaranteed to establish its statement, without any need to read the proof itself.
Its community library, Mathlib, provides reusable mathematical definitions
and theorems~\cite{mathlib2020}. At the start of our project, however,
Mathlib lacked much of the reusable formal infrastructure required for the
geometric analysis in the Poincar\'e proof. Outside Mathlib, relevant formal developments were also limited and dispersed across individual projects. An early contribution during our preparatory phase was Chow et al.'s formalization of Hamilton's three-manifold theorem~\cite{chowLiaoQin2026}.\footnote{The public Lean source tree records selected adaptations from \emph{Differential Geometry in Lean 4}~\cite{differentialGeometry2026}, including its adapted De Giorgi material~\cite{armstrongKempe2026,degiorgiRepo2026}, as well as from ClassificationOfSurfaces~\cite{classificationSurfaces2026}, Tau Ceti~\cite{tauCeti2026}, and Mathlib~\cite{mathlib2020}. The fixed upstream revisions and file-level scopes are documented in the repository's \href{https://github.com/frenzymath/PoincareConjecture/blob/62db01af9388dc1d9b56b66d6674f4c09b2318a2/MODIFICATIONS.md}{\texttt{MODIFICATIONS.md}}.} Our Poincar\'e formalization was developed independently of the contemporaneous work of Qin et al.~\cite{qinLiaoKhaitanChow2026}. The developments use different Ricci-flow interfaces and final proof assemblies; ours follows separately designed and reviewed milestones based on Morgan--Tian.

\paragraph{The proof of the Poincar\'e conjecture.} The proof follows the Ricci flow program initiated
by Hamilton~\cite{Hamilton3MPRC}. Perelman's three preprints in 2002--2003
provided the key arguments in a highly condensed
form~\cite{PerelmanI,PerelmanII,PerelmanIII}. Subsequent accounts expanded
these arguments in detail~\cite{KleinerLott2008,CaoZhu2006,MT2007,MT2015Correction}.
Our formalization follows Morgan--Tian's 2007 monograph, which develops the
mathematical background and the proof of the conjecture.

\paragraph{Prior work on autoformalization.}
Early work on autoformalization studied the translation of informal
mathematical statements into formal languages
~\cite{wang2018neural,wu2022autoformalization}. Subsequent research developed
datasets and methods for statement formalization in Lean
~\cite{jiang2023mma,gao2025herald,liu2025atlas,wang2026aria}. In parallel,
language models advanced proof generation for given formal goals
~\cite{polu2020gptf,yang2023leandojo,ren2025deepseek,hubert2026alphaproof}.
These directions also interact: DSP showed how natural language proofs could guide formal proof construction
~\cite{jiang2023dsp}, while subsequent work studied the translation of both
statements and proofs into Lean~\cite{jana2026proofbridge}. Agent systems
build on these capabilities, using library retrieval and proof-assistant
feedback to carry out and revise sequences of proof tasks
~\cite{thakur2024copra,liu2026numina,ju2026archon}. Recent projects have also
extended autoformalization to mathematical developments containing many
interdependent results~\cite{urban2026topology,wang2026m2f,gloeckle2026textbook}.
In September 2026,
the formalization of Fermat's Last Theorem using Prove2Me produced a complete
computer-checked proof. Its published account emphasizes the
resulting development while giving less detail about the division of human and
AI work~\cite{prove2me2026,anthropicFLT2026}. The challenge
addressed here is to extend this approach to the Poincar\'e proof while
developing much of the geometric analysis absent from existing reusable
libraries.

\section{Methodology}
\label{sec:methodology}

Our first attempts to formalize material related to the Poincar\'e conjecture
began in May 2026. In September, we redesigned the workflow, restarted the
formalization from scratch, and completed the main development in slightly
more than two weeks.

During the initial phase, we collected books and papers on Riemannian geometry \cite{doCarmo,Petersen,Gallot,Sakai,CheegerEbin}, algebraic topology
\cite{HatcherAT2002}, PDEs \cite{Evans2010}, three-manifold topology
\cite{SchoenYau,MT2007}, and Ricci flow
\cite{ChowKnopf,ChowLuNi,Hamilton3MPRC,Hamiltonharnack,Shi1,Shi2,
PerelmanI,PerelmanII,PerelmanIII,CaoZhu2006}. We initially planned to
formalize substantial portions of this background before addressing the
Poincar\'e conjecture itself. This approach progressed slowly, and the resulting
Lean code did not meet our expectations for quality or organization. The
reasons for this difference in performance are discussed in
Section~\ref{sec:acceleration}.

By late July, we had completed a Lean formalization of the Hopf--Rinow
theorem and reviewed its mathematical correctness. After an initial attempt
with an AI system failed, a mathematician reorganized the proof into a clearer
route, enabling the system to complete the development. Throughout August,
we continued formalizing foundational material from these sources and
completed a natural-language blueprint of the Poincar\'e proof following
Morgan--Tian.

\begin{figure}[htbp]
\centering
\includegraphics[width=0.8\textwidth]{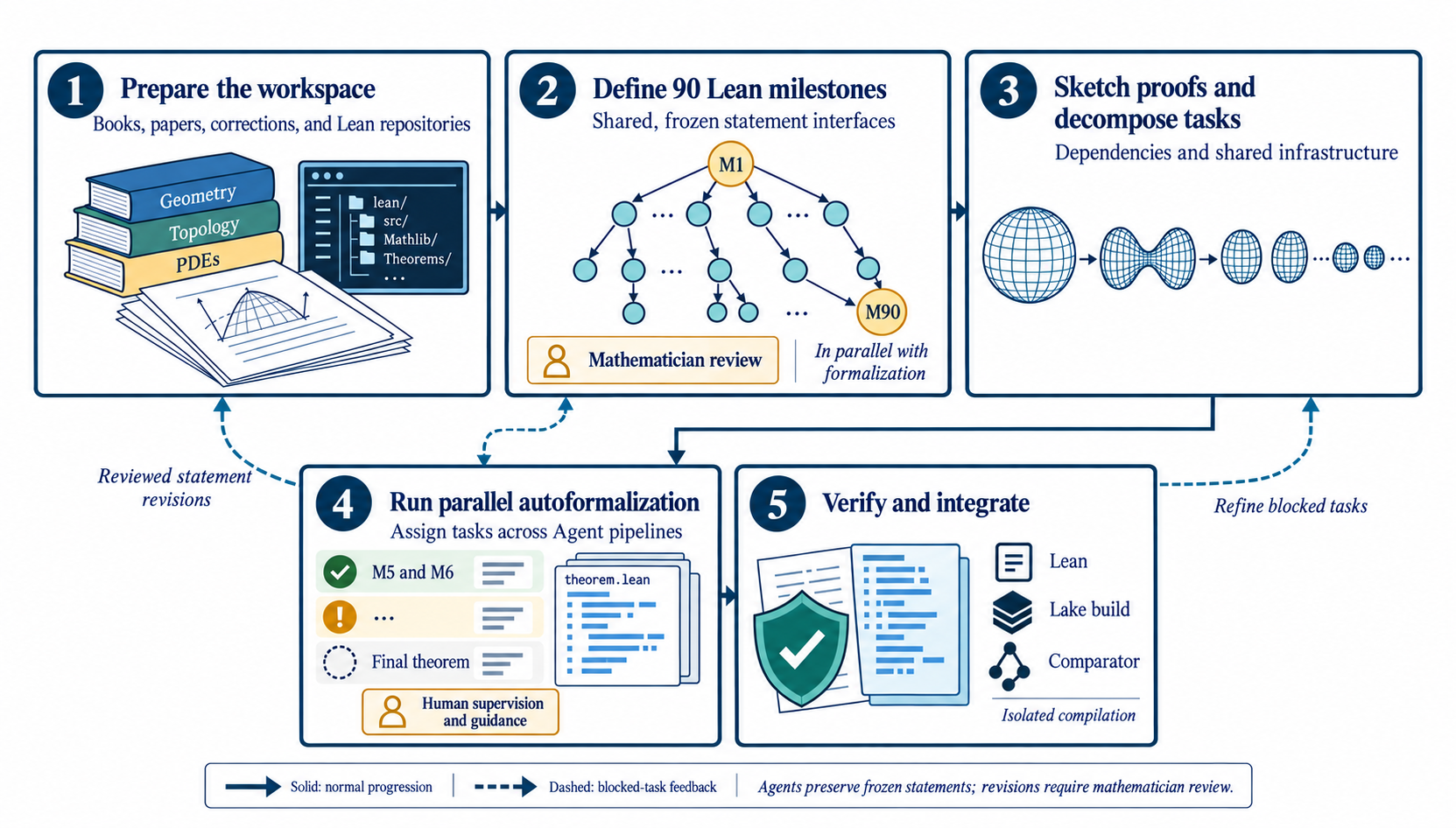}
\caption{
Workflow for converting a mathematical target into a verified Lean
formalization. Solid arrows indicate the ordinary progression of work.
Dashed arrows indicate that a blocked task can return to source gathering,
statement revision, or further decomposition. Human decisions determine the
mathematical scope, review priorities, and acceptance criteria, while agents
assist with source analysis, formalization, monitoring, and diagnosis.
}
\label{fig:workflow}
\end{figure}

Drawing on this preparation, we redesigned the workflow summarized in
Figure~\ref{fig:workflow}. We collected the relevant sources, corrections, and repositories, and
organized them into a proof skeleton following the Morgan--Tian exposition
\cite{MT2007}. This skeleton consisted of 90 milestones distributed throughout
the intended argument. Each milestone included a Lean statement and the
definitions required by that statement. The resulting milestone interfaces
were then used for parallel work across independent tasks.

The milestone statements were reviewed by mathematicians in parallel with the formalization work. This review established a common standard for the principal results and helped prevent the growing codebase from diverging from the intended mathematics. Since it was not possible to review every line of agent-generated code, the milestone statements provided checkpoints at which the mathematical content could be examined independently of the eventual
implementation.

The milestone structure also exposed the dependency chains in the proof. For
each milestone, agents were asked to use the mathematical references to produce an informal proof sketch and to identify the results on which the milestone depended. These sketches made dependencies visible and helped group related milestones into larger tasks, for example when several milestones required the same underlying infrastructure. Because the project used several autoformalization pipelines, the reviewed milestone statements also served as
common interfaces between those pipelines.

Some tasks required direct human supervision, often simply by bypassing an agent's self-imposed constraints or correcting a subtle logical gap. Interactive sessions with coding
agents such as Codex and Claude Code were used to monitor progress, identify unproductive proof attempts, and diagnose blockers. In many cases, a discussion with the agent was enough to unblock a task. In other cases, supervision exposed an inconsistency in a frozen Lean statement. Such statements were not changed by the agents; the issue was instead reviewed by the relevant mathematicians and corrected at the coordination level.

The human role therefore remained focused on mathematical scope, statement review, task prioritization, and the resolution of exceptional difficulties.
This division of labor made it possible to supervise the development at a high level without inspecting every agent action or every line of Lean code. Once a task satisfied its milestone
interface, its changes could be merged into the shared repository. This experience showed that the organization of the
formalization process was as important as the capabilities of the underlying
agents.

\subsection{Formal Statement Review}
\label{sec:statements}

Statement review was the checkpoint at which we assessed the mathematical target independently of its proof. For each milestone, reviewers examined whether the definitions, hypotheses, conclusion, and intended dependencies expressed the result required by the proof route. The final theorem statements were kept separate from the supporting proof modules, so that a reviewer could evaluate what Lean was asked to prove without reconstructing an agent's proof history.

Each statement was accompanied by its mathematical source, relevant corrections, dependencies, and intended level of generality. The review record connected these choices to the corresponding Lean declaration and recorded later revisions. This made statement review a distinct and auditable stage, rather than an inference from successful compilation.

Reviewers compared the natural-language statement with the fixed version of Mathlib and with the requirements of later milestones. External formalizations could suggest definitions or proof interfaces, but any adaptation was checked against the source mathematics and the project's chosen meanings. Keeping the explanation beside the Lean statement made this comparison direct.

The review had two passes. The first checked the written definitions and statement against the mathematical sources and downstream uses. The second inspected the declaration produced by Lean's elaborator, including information left implicit in the source code, to detect unintended parameters or assumptions. Statement types were also required to be independent of admitted proofs. These checks validated the target as a basis for proof construction; they did not by themselves prove the theorem.

Mathematicians participated from the statement-design stage and resolved ambiguities before they propagated into dependent tasks. When review found that a shared statement needed correction, the coordinator recorded the revised basis and reopened the affected reviews; unaffected milestones could continue. Human intervention therefore controlled the mathematical scope while preserving the distinction between a reviewed statement and a mechanically checked proof.

\subsection{Milestone Decomposition and Dependency Management}
\label{sec:milestones}

Milestone design was a joint mathematical and engineering task. Mathematicians selected a coherent proof route, supplied arguments and references, and identified intermediate results that could be reviewed independently. Engineers and Lean experts assessed the available formal infrastructure and the work needed to connect those results in Lean. Named theorems in the Morgan--Tian exposition provided initial landmarks; a usable formalization blueprint also had to specify the definitions, supporting constructions, and dependencies between them. The milestones therefore reflected both the logical structure of the mathematics and the practical organization of its implementation.

In choosing the spacing between milestones, we considered the expected amount of Lean code and prerequisite development. A short deduction on paper may require substantial formal work when suitable definitions, estimates, or links between representations are missing. Such gaps warranted additional intermediate tasks. Conversely, closely related results could share a task when they depended on the same construction. This assessment concerned the work needed to produce a useful, reviewable result, rather than a fixed number of textbook pages or lines of code.

The appropriate level of generality was also a design choice. To simplify formalization, a result could be specialized to dimension three, or restricted to a weaker conclusion, when that was sufficient for every subsequent use. Other milestones benefited from stronger conclusions: returning a constructed object together with the properties needed later could avoid repeated constructions and proofs of compatibility. The annulus and deformation stages illustrate this choice in Section~\ref{sec:bottlenecks}. Such revisions required mathematical review to ensure that the final theorem remained supported and that additional conclusions were proved from justified assumptions. Agents could report a difficulty, but changes to an agreed target were coordinated explicitly.

Each milestone specified its definitions, assumptions, and required outputs before proof assignment. A directed dependency graph recorded which earlier constructions or results each task would use. Shared foundations were assigned to a single owner, and independent tasks were developed in separate workspaces using agreed definitions and statements. As implementation exposed missing prerequisites or unsuitable task boundaries, we revised the graph and the affected assignments while allowing unaffected work to continue.

This structure made expert guidance useful throughout execution. Mathematicians with a high-level view of the proof and relevant formalization-engineering experience could often identify a missing argument, an unsuitable target, or a dependency problem early, and give effective guidance before extensive code accumulated. Engineers could then translate that diagnosis into revised tasks and dependencies. The combination of mathematical judgment, Lean implementation experience, and automated proof construction therefore provided a practical basis for coordinating and reviewing a development too large to inspect line by line.

\subsection{Agent Execution and Team Coordination}
\label{sec:execution}
\begin{figure}[htbp]
    \centering
    \includegraphics[width=0.8\textwidth]{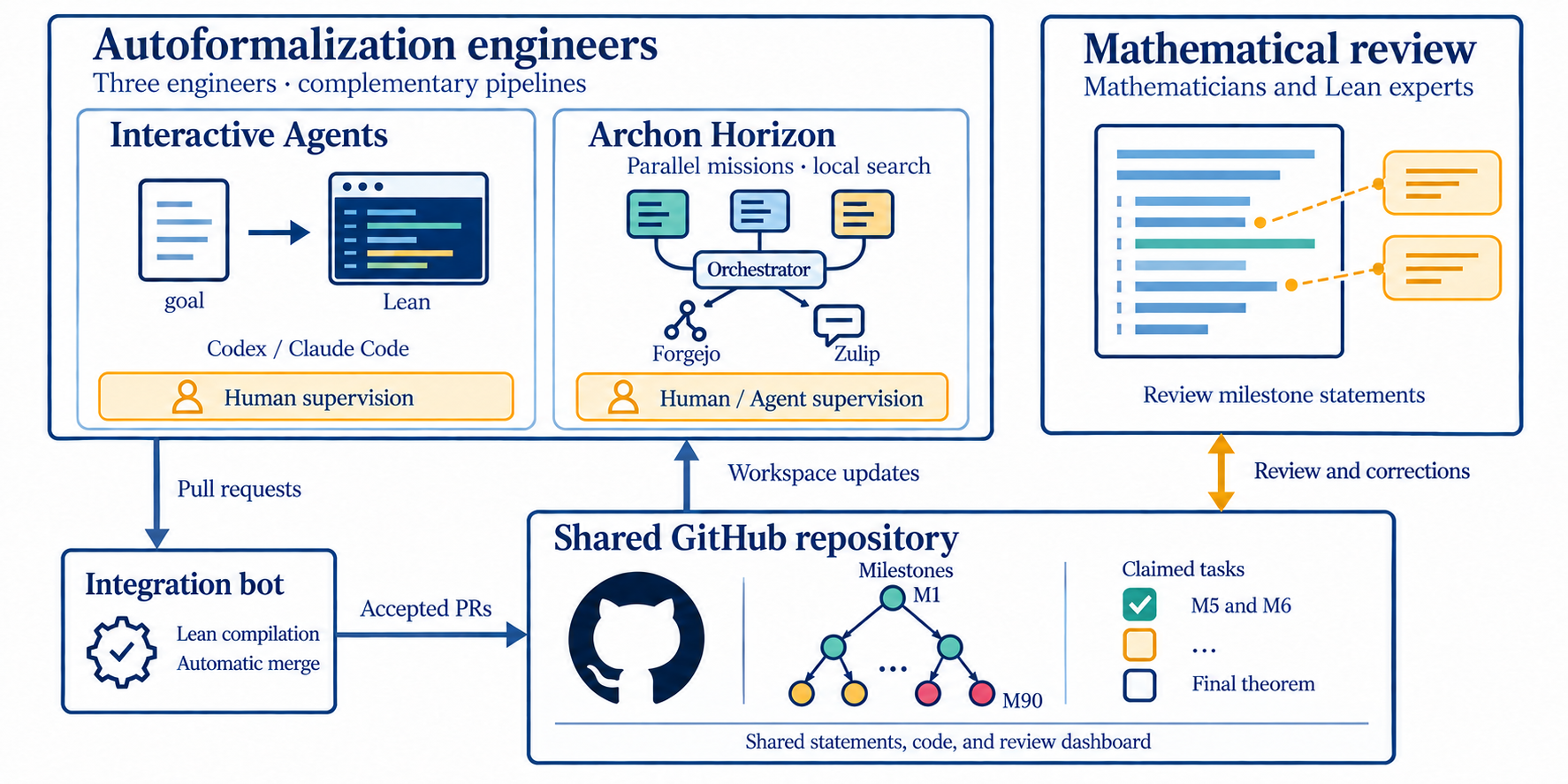}
    \caption{
        Team organization and coordination for autoformalization. Engineers use
        goal-oriented interactive coding agents and Archon Horizon to develop
        Lean proofs under human supervision. Contributions pass through an
        integration bot that compiles Lean code and automatically merges
        accepted pull requests into the shared GitHub repository. The repository
        coordinates milestone claims, distributes updates to individual
        workspaces, and supports statement review and corrections by
        mathematicians and Lean experts.
    }
    \label{fig:team-coordination}
\end{figure}

The formalization of the Poincar\'e conjecture involved three AI engineers, together with mathematicians and Lean experts. As illustrated in Figure~\ref{fig:team-coordination}, each engineer used a distinct combination of models, agent frameworks, and computing resources. This diversity made it possible to use complementary methods of autoformalization, but also required coordination to avoid duplicated work, incompatible interfaces, and conflicting versions of the same result.

For different scales and failure modes, we primarily used two complementary modes: interactive coding-agent sessions for tightly scoped, high-context interventions, and a distributed mission workflow for proof branches that could be decomposed and run in parallel. The first mode used interactive sessions, principally with Codex and Claude Code. Each session was organized around a proof-branch harness rather than an isolated coding prompt. Before an agent began, an engineer fixed the branch boundary, milestone endpoint, definitions and interfaces to be used, predecessor and consumer dependencies, relevant mathematical sources, expected artifacts, and compilation and admission checks. The engineer then used the session to search and filter the pinned Mathlib APIs and source references, translate the selected argument, monitor the resulting declarations, and revise the route when a mathematical or implementation obstruction appeared. This architecture kept local proof search tied to the surrounding dependency graph. GPT-6-Astra carried most of the long-context statement formalization and proof construction, while Fable-5.1 was used for independent source-alignment checks, diagnosis of blocked branches, and refinement of the proposed harness. This division of labor proved effective in practice: one model could sustain the main construction while the other supplied targeted scrutiny and alternative directions.

The second method used Archon Horizon.\footnote{\url{https://github.com/frenzymath/Archon-Horizon}}
Archon Horizon is a distributed control plane for formalization projects. It
represents a project as a collection of bounded missions with explicit goals,
dependencies, and completion conditions, rather than as one long interactive
session. Its informal project representation is a directed acyclic graph of
Markdown documents. These documents can contain Lean snippets, links to
commits and sources, informal proofs, and other task-specific material.

This structure supports parallel execution. Horizon can dispatch independent
missions to agents running on different configured servers, while shared
repositories and communication channels maintain a common view of the project.
This reduces local computing bottlenecks and allows several formalization tasks
to progress concurrently. Engineers can still supervise the system through
interactive agent sessions: they can inspect progress, supply context, launch
or cancel missions, and intervene when a task appears blocked. Horizon
therefore combines distributed execution with the possibility of targeted human
oversight.

Because the engineers used different methods and servers, coordination required
a shared repository. Contributors claimed milestones there and submitted
completed work through pull requests. An integration bot independently compiled
the Lean code and automatically merged accepted changes. When a peer completed
a milestone, the other engineers could incorporate the resulting changes into
their own workspaces and continue from the updated development.

The shared repository also supported a review dashboard. Mathematical reviewers
could inspect and comment on the formal statement of each milestone. When a
review identified an issue, an engineer revised the statement in the main
repository before later tasks relied on it. This made review status visible to
all participants and allowed corrections to propagate quickly through the
formalization workflow.

\subsection{Project Resources}
\label{sec:timeline-resources}
We now describe the resources behind the project: the models used, the division of labor among them, and the resulting cost. GPT-6-Astra formalized the milestone statements in Lean, and both GPT-6-Astra
and Fable-5.1 performed multiple rounds of statement review. Proof construction
relied primarily on GPT-6-Astra, with occasional consultation of Fable-5.1.

The cost of AI subscriptions and rented servers was approximately \$25{,}000.
About \$20{,}000 went to AI subscriptions, primarily GPT Pro $20\times$ plans,
roughly equivalent to 100 subscriptions at \$200 each. This subscription
equivalent provides only a rough reference for model usage, given quota resets
and fluctuations in OpenAI's usage allowances. The remaining approximately
\$5{,}000 covered the rental of Linux servers, used mainly for Lean compilation.

\section{Analysis}
\label{sec:analysis}

\subsection{Agent Behavior and Human Intervention}
\label{sec:agent-analysis}

Agents typically worked through repeated cycles of source and library search, informal reasoning, Lean implementation, and compilation. They assisted with translating arguments, connecting existing results, repairing local proofs, and checking code mechanically. Recurring difficulties arose when a task required judging whether a stronger claim followed from the reference, whether two representations described the same geometric object, or whether an intermediate result would help prove the final theorem. These decisions required review beyond checking whether individual code fragments compiled.

When progress stalled, we found that the difficulties mainly fell into three categories, which helped us address each one at its source:
\begin{itemize}
    \item \textbf{Mathematical gaps.} These required a more detailed argument, a suitable reference, or a reviewed correction to the statement or its hypotheses.
    \item \textbf{Lean implementation problems.} These called for a better representation, an existing library result, or a lemma relating two definitions.
    \item \textbf{Coordination problems.} These included duplicated work, unstable ownership, or incompatible versions of a shared statement, and required changes to assignments or dependencies.
\end{itemize}
This diagnosis determined whether further proof search was useful and what information an agent needed to proceed.

Engineers used the current dependency graph to assign independent milestones whose statements were sufficiently stable. They monitored ongoing attempts and adjusted the assignments as missing prerequisites became apparent. Mathematical review helped identify which intermediate goals supported the intended argument, allowing unproductive directions to be curtailed while other tasks continued.

Failed attempts also supplied diagnostic evidence. A repeatedly unsuccessful
task could be too broad, rely on an unavailable assumption, or require
information omitted from an earlier result. Task descriptions, review notes,
and progress reports preserved these issues for targeted discussion. Human
intervention therefore directed the work as well as reviewing its output:
mathematicians clarified the mathematical requirements, engineers organized
their implementation, and agents carried out the resulting proof tasks.

\subsection{Mathematical Bottlenecks and Formalization Experience}
\label{sec:bottlenecks}

Mathematicians contributed to the design and progress of the formalization through choices about scope, proof strategy, and the information passed between tasks. These interventions fall into three recurring classes. First, mathematicians set the scope of an interface: they decide which hypotheses, conclusions, and level of generality are mathematically justified and useful to later milestones. Second, they diagnose proof-route and representation gaps: they identify missing intermediate arguments, select the relevant theorem or local geometric model, and separate a mathematical obstruction from a Lean encoding problem. Third, they manage interfaces and downstream reuse: they expose hidden assumptions, decide when an output should be strengthened for later consumers, and keep dependencies and chronology explicit. The examples below follow this order: M22 concerns scope, M28 concerns a missing proof route, and M64--M65 concern interface correction and reusable output.

The first concerned the scope of intermediate results. In the noncollapsing argument (M22), the uniform estimate for the non-round solutions under consideration was formulated in dimension three, while an auxiliary result about asymptotic volume ratios was retained in general dimension to support induction on dimension. These choices required understanding how each result was proved and used later. Restricting a statement can remove unnecessary work, but retaining generality can also provide the structure needed by the proof itself.

The second concerned a curvature estimate used to analyze singularities of Ricci flow (M28). Its proof required a local limiting construction that was not supplied by the available compactness theorem for complete spaces. Mathematical review identified the missing local compactness argument and the geometric contradiction needed to finish the proof, with references to the relevant source arguments. This turned an apparent failure to apply an existing theorem into explicit supporting tasks, giving agents a clearer route through the analytical difficulty.

The third concerned comparing and deforming families of curves (M64--M65). Review of the annulus comparison required an explicit initial annulus connecting the boundary loops, which subsequent applications had to supply: equal winding around an auxiliary circle did not ensure that such a connection existed. The task also needed more than a statement that a suitable approximation existed. Its output retained a family of polygonal approximations, their error and area bounds, and estimates on the associated evolving curves. The next milestone could then use those same objects and estimates. Mathematical review thus both corrected a missing assumption and identified stronger outputs that made the subsequent formalization easier to organize.

These interventions made expert input actionable: a revised statement, a more detailed proof route, or a reusable construction could guide many subsequent agent steps. The value of mathematical participation lay in identifying what should be formalized and why it was sufficient, while Lean expertise and automated proof construction made those decisions executable and checkable.

\subsection{Accelerating the Formalization}
\label{sec:acceleration}
After redesigning the workflow in September as described in Section~\ref{sec:methodology}, we completed the main
formalization phase in slightly more than two weeks. This acceleration did not
arise simply from asking agents to work faster or from allocating more compute: the earlier project had already produced roughly 500{,}000 comment-free lines
of book-based Lean source, as well as a blueprint and tracked many individual nodes.
This earlier source collection was separate from the later full development,
whose dependency cone contains about $2.8$ million of the original $3.2$
million lines.
The central change was to turn a comparatively small collection of reviewed
milestone statements into fixed interfaces for the entire project.

Earlier tasks were often broad, for example formalizing a chapter, a section
of the blueprint, or a node in a dependency graph. Such tasks could produce
substantial verified progress while still lacking a precise and independently
checkable endpoint, making progress difficult to review and the causes of
blockers difficult to identify. In the redesigned workflow, the proof was organized around
90 milestone statements distributed throughout the intended argument. Each
statement specified the assumptions and conclusion to be preserved, its
dependencies, and the condition under which the corresponding task could be
considered complete. The milestone therefore provided a concrete location at
which progress could be inspected and the mathematical content reviewed.

The timing of review also changed, mitigating the human-review bottleneck. In the earlier attempts, Lean code was
reviewed after it had been formalized, making it difficult for human feedback to shape the tasks assigned to the autoformalization system. In the new workflow, mathematicians
reviewed the central milestone statements before, or in parallel with, their
formalization. These reviewed statements constrained subsequent agent work and
reduced the risk that many locally successful proofs would accumulate around
an unsuitable definition or an incorrect formulation of a major result.

The new structure made both progress and blockers easier to understand. For
example, the earlier Ricci-flow work encountered separate difficulties in
curvature evolution, maximum-principle arguments, derivative estimates, and
endpoint regularity. The milestone decomposition made interfaces between these
arguments explicit, instead of leaving them as unresolved parts of a single
large analytic task. We could then identify whether a blocked task
required a missing lemma, a more detailed informal argument, an additional
reference, further decomposition, or a revision of the statement.

Frozen milestone statements also made parallel work practical. They were
particularly well suited to a framework such as Archon Horizon, which assigns
bounded tasks to different agents and servers. Different autoformalization
pipelines could work on independent dependency chains while sharing the same
definitions and theorem interfaces. Once a task satisfied its milestone
interface and compiled in the shared repository, it could be integrated without
requiring the other engineers to reconstruct its complete proof history.

In our project records, the milestone structure was associated with three operational changes: it constrained the
formalization around reviewed mathematical targets, localized failures, and
provided a reliable measure of progress. The earlier work nevertheless
remained valuable, since it supplied mathematical sources, initial blueprint
material, Lean experience, and a clearer understanding of the forms of
decomposition that were required for the final workflow.

\subsection{Comparison with FLT and Reusability}
\label{sec:flt}

The recent Anthropic formalization of Fermat's Last Theorem (FLT) provides a
useful comparison because both projects produced an end-to-end Lean
verification of a major theorem through coordinated agent work. The two
projects began from different foundations, however. Anthropic's development \cite{anthropicFLT2026}
built on Mathlib \cite{mathlib2020}, the ongoing Imperial College FLT project \cite{imperialFLT}, and the existing
formalization of FLT for regular primes
\cite{fltRegular}. By contrast, the
Poincar\'e project required the construction of project-specific infrastructure
for substantial parts of Riemannian geometry, Ricci flow, geometric limits,
surgery, and three-manifold topology before the final argument could be
formalized.

The projects nevertheless share an important organizational principle.
Anthropic reports that early FLT attempts failed in part because agents lost
track of the evolving project state. Their successful workflow used Prove2Me \cite{prove2me2026}
to maintain a directed acyclic graph of statements, separate statements from
proofs, and support search and reuse through descriptions of the available
results \cite{anthropicFLT2026,prove2me2026}. Our workflow similarly used
explicit dependencies and fixed statement interfaces to coordinate parallel
proof efforts. The main difference is that our milestones were selected in
advance from the Morgan--Tian route and reviewed by mathematicians before, or
while, formalization proceeded. These reviewed statements served as stable
interfaces between several independent pipelines. This suggests that successful autoformalization depends not only on stronger
agents or greater parallelism. Without a sufficiently structured workflow,
either approach can eventually reach a point at which accumulated inconsistencies,
unresolved dependencies, or loss of project state make a restart necessary, as
occurred in the early attempts of both projects.
The organization of the project appears to have contributed to this
result: careful preparation before formalization, followed by explicit task
decomposition, stable interfaces, and coordination throughout execution.

FLT also makes clear the distinction between verifying a final theorem and
developing reusable mathematical infrastructure. The released FLT repository
describes itself as a research artifact that is not maintained and does not
accept contributions \cite{anthropicFLTArtifact2026}. Its successful
verification does not imply that its definitions, module structure, and proof
interfaces are immediately suitable for reuse in later Lean developments. This
does not prevent later projects from studying or extracting individual parts of
the artifact, but substantial reorganization and mathematical design work may
be required before those parts function as a maintained library.

We face a related distinction in the present project. Mathematical review of
the central milestone statements was intended both to secure the final theorem
and to improve the interfaces around which the code was built. Nevertheless,
we do not regard the current development as finished. Its initial decomposition
follows the Morgan--Tian exposition \cite{MT2007}, but the Lean proofs
produced by the agents do not always follow the same route as the reference.
Analyzing these proof paths, especially around tasks that initially
blocked the agents, is part of our future work: they may contain useful alternative
arguments, more effective intermediate statements, or mathematical
observations that deserve to be recorded independently of the formalization.

We also aim to improve the development as Lean code. A dependency-cone
extraction reduced the initial approximately $3.2$ million-line development to
approximately $2.8$ million lines while preserving Comparator verification of
the final theorem. This is only a first step. We plan to use our internal
analysis and refactoring tools to remove duplication, improve proof structure
and documentation, generalize definitions and theorem statements where
appropriate, and reorganize the project into a reusable geometric-analysis
library. The present formalization is therefore both a verified proof and a
checked source from which a more maintainable mathematical library can be
developed. Future refactoring may make selected components suitable
for direct contribution to maintained libraries such as TauCeti
~\cite{tauCeti2026}.

\section{Conclusion and Outlook}
\label{sec:conclusion}
Using publicly available models, we formalized the Poincar\'e conjecture
together with much of its prerequisite theory. For this project, several months of preparation supplied the proof
blueprint and informed the milestone structure. Preparing the milestones
required organizing the proof route, dividing it into units of manageable
formalization size, and making their dependencies explicit.
This structure made parallel work possible and helped us locate problems that
required mathematical judgment. Experts provided high-level guidance when
statements needed revision or proof attempts stalled. This experience shows
that an independent research team can achieve large-scale formal verification
with human--AI collaboration. As models improve, we expect them to take on
more of the mathematical preparation and substantially shorten the time
required by the formalization process.

In the public snapshot, about $48\%$ of the Lean source is devoted to
background theory and shared mathematical infrastructure.\footnote{This
classification is based on GPT-6-Astra's assessment, which we prompted
to distinguish source devoted to the Perelman proof from background and
supporting theory. The classification refers to commit~\href{https://github.com/frenzymath/PoincareConjecture/commit/3d7318c32025ccb30465dd29850e95a9e2aa1526}{\texttt{3d7318c}}. Examples include Bishop--Gromov volume comparison,
short-time existence for Ricci flow, Shi's derivative estimates, the Hurewicz
theorem, and the smooth Schoenflies theorem in dimension three;
representative declarations are listed in
Appendix~\ref{app:intermediate-formalizations}.} For example, our development
of smooth structures on compact connected topological three-manifolds spans
about $650{,}000$ lines of Lean, including supporting theory, although
Morgan--Tian invokes this classical result in an opening footnote.
Reliable, reusable mathematical infrastructure is central to reducing the
difficulty and cost of formalization. The key is to provide carefully reviewed
definitions and theorem statements that accurately express the intended
mathematics and can be reused across projects. Such libraries can reduce the
need to rebuild foundational theory in every project. An important direction
for future work is to turn the material from this project, including
its many intermediate formalizations of results, into a
maintained geometric-analysis library. This requires reviewing its mathematical
interfaces, removing duplication, and improving its organization and
documentation. Its success should be judged by how much work and cost it saves
in later formalizations. Meanwhile, the proof blueprint can be refined
into a clearer mathematical account of the Poincar\'e argument, making its
dependencies explicit and clarifying the role of its intermediate
conclusions.

As proof generation and formal verification become cheaper, we hope that
affordable tools will let an independent mathematician easily use formalization
to verify their own research: stating goals, preparing milestones, guiding proof
construction, reviewing formal statements, and finally obtaining mechanically
verified results. At a broader level, formalization can also help clarify and
improve mathematics itself. In a separate Laver-function formalization project,
our formalization identified corrections and clarifications that were
incorporated into a revised paper~\cite{laverFormalization2026}. For the
Poincar\'e conjecture, however, much remains to be done to re-examine its proof
mathematically with the help of formalization. The formalization has
also revealed potential for mathematical revisions that could correct or
simplify parts of the argument and deepen our understanding of its structure.
We believe that formalization,
if done properly, can help mathematicians reorganize mathematics by clarifying
definitions, making dependencies explicit, and revealing connections between
arguments. To make this capability accessible to the mathematical community
more broadly, we invite mathematicians to carry their high-level mathematical
knowledge into human--AI interactions centered on Lean, and to help build the
libraries, tools, and practices this development requires.

\section*{Acknowledgments}

The authors would like to warmly thank Wangjian Jian, Xilun Li, Zhengnan Chen for providing expert mathematical consultation throughout the formalization of the Poincar\'e conjecture, and Zekun Sheng, Nan Wu, Wanxu Yang, Hongyu Chen for reviewing the statement of the final Poincar\'e theorem and those of the key intermediate milestones, helping ensure that the formalization targeted the intended mathematics.

This work is supported in part by the Fundamental and Interdisciplinary Disciplines Breakthrough Plan of the Ministry of Education of China (JYB2025XDXM113), the National Key R\&D Program of China (grant 2024YFA1014000), and the New Cornerstone Investigator Program.

\bibliographystyle{plainnat}
\bibliography{references}

\clearpage
\beginappendix
\section[Main Results in Lean]{Main Results in Lean}
\label{app:main-results}

The following Lean definitions state the smooth and topological
versions of the Poincar\'e theorem using only Mathlib.\footnote{Source:
\url{https://github.com/frenzymath/PoincareConjecture/blob/3d7318c32025ccb30465dd29850e95a9e2aa1526/Comparator/Challenge.lean}.}

\begin{lstlisting}[style=leanresult]
import Mathlib

set_option autoImplicit false
open scoped Manifold ContDiff
universe u

namespace PoincareConjecture

abbrev ThreeSphere := Metric.sphere (0 : EuclideanSpace ℝ (Fin 4)) 1

def SmoothPoincare : Prop :=
  ∀ (M : Type u) [TopologicalSpace M] [T2Space M] [SecondCountableTopology M]
    [ChartedSpace (EuclideanSpace ℝ (Fin 3)) M] [IsManifold ($\mathcal{R}$ 3) ∞ M]
    [CompactSpace M] [SimplyConnectedSpace M],
    Nonempty (Diffeomorph ($\mathcal{R}$ 3) ($\mathcal{R}$ 3) M ThreeSphere ∞)

def TopologicalPoincare : Prop :=
  ∀ (M : Type u) [TopologicalSpace M] [T2Space M] [SecondCountableTopology M]
    [ChartedSpace (EuclideanSpace ℝ (Fin 3)) M]
    [CompactSpace M] [SimplyConnectedSpace M], Nonempty (M ≃ₜ ThreeSphere)

end PoincareConjecture
\end{lstlisting}

\subsection{Representative Intermediate Lean Formalizations}
\label{app:intermediate-formalizations}

The table lists representative intermediate Lean declarations
produced during the project. It documents the scope of the automated
development; it does not establish that the corresponding informal statements
have been mathematically validated or that the declarations are ready for
reuse.\footnote{Source snapshot:
\url{https://github.com/frenzymath/PoincareConjecture/tree/62db01af9388dc1d9b56b66d6674f4c09b2318a2}.}

\begin{center}
\scriptsize
\renewcommand{\arraystretch}{1.2}
\begin{tabular}{@{}p{0.32\textwidth} p{0.33\textwidth} p{0.29\textwidth}@{}}
\hline
\textbf{Intended mathematical topic} & \textbf{Lean declaration} & \textbf{File} \\
\hline
Short-time Ricci-flow existence on compact smooth manifolds in arbitrary
dimension
& \texttt{PoincareMT.}\allowbreak\texttt{shortTimeRicciFlowExistence}
& \href{https://github.com/frenzymath/PoincareConjecture/blob/62db01af9388dc1d9b56b66d6674f4c09b2318a2/PoincareLib/Geometry/RicciFlow/Local/ShortTime.lean}{\nolinkurl{PoincareLib/Geometry/RicciFlow/Local/ShortTime.lean}} \\
Bishop--Gromov volume comparison for complete connected manifolds with
nonnegative Ricci curvature
& \texttt{PoincareMT.}\allowbreak\texttt{asymptoticVolumeRatio}\allowbreak\texttt{BishopGromov}
& \href{https://github.com/frenzymath/PoincareConjecture/blob/62db01af9388dc1d9b56b66d6674f4c09b2318a2/PoincareLib/Geometry/RicciFlow/AncientKappa/Volume/BishopGromov.lean}{\nolinkurl{PoincareLib/Geometry/RicciFlow/AncientKappa/Volume/BishopGromov.lean}} \\
Local Shi derivative estimates for Ricci flows
& \texttt{PoincareMT.}\allowbreak\texttt{local\_curvatureDerivative}\allowbreak\texttt{\_bound}
& \href{https://github.com/frenzymath/PoincareConjecture/blob/62db01af9388dc1d9b56b66d6674f4c09b2318a2/PoincareLib/Geometry/RicciFlow/Curvature/Estimates/Derivative.lean}{\nolinkurl{PoincareLib/Geometry/RicciFlow/Curvature/Estimates/Derivative.lean}} \\
Integral Hurewicz bridge used in the three-manifold argument
& \texttt{Poincare.Topology.}\allowbreak\texttt{exists\_homotopyGroupPi}\allowbreak\texttt{\_mulEquiv\_of\_integral}\allowbreak\texttt{\_hurewicz\_bijective}
& \href{https://github.com/frenzymath/PoincareConjecture/blob/62db01af9388dc1d9b56b66d6674f4c09b2318a2/PoincareLib/Topology/Homotopy/Hurewicz/HurewiczAlgebra.lean}{\nolinkurl{PoincareLib/Topology/Homotopy/Hurewicz/HurewiczAlgebra.lean}} \\
Smooth three-dimensional Schoenflies neighborhood theorem for a smoothly
embedded two-sphere in Euclidean three-space
& \texttt{Poincare.Manifold.SmoothDomain.}\allowbreak\texttt{exists\_ball\_neighborhood}
& \href{https://github.com/frenzymath/PoincareConjecture/blob/62db01af9388dc1d9b56b66d6674f4c09b2318a2/PoincareLib/Topology/Manifold/Schoenflies/Neighborhood.lean}{\nolinkurl{PoincareLib/Topology/Manifold/Schoenflies/Neighborhood.lean}} \\
\hline
\end{tabular}
\end{center}
\end{document}